\pdfoutput=1

\documentclass[11pt]{article}

\usepackage[preprint]{acl}

\usepackage{times}
\usepackage{latexsym}
\usepackage{amsmath}
\usepackage{amssymb}
\usepackage{booktabs}
\usepackage{multirow}
\usepackage{multicol}
\usepackage{subcaption}
\usepackage{makecell}
\usepackage{xcolor}
\usepackage{arydshln}

\usepackage[T1]{fontenc}

\usepackage[utf8]{inputenc}

\usepackage{microtype}

\usepackage{inconsolata}

\usepackage{graphicx}

\title{Smoothing Grounding and Reasoning for MLLM-Powered \\ GUI Agents with Query-Oriented Pivot Tasks}

\author{Zongru Wu, Pengzhou Cheng, Zheng Wu, Tianjie Ju, \\ \textbf{Zhuosheng Zhang}$^{\textcolor{darkblue}{*}}$, \textbf{Gongshen Liu}\thanks{Corresponding authors.} \\
School of Electronic Information and Electrical Engineering, \\ Shanghai Jiao Tong University \\
  \texttt{\{wuzongru,cpztsm520,wzh815918208,jometeorie,zhangzs,lgshen\}@sjtu.edu.cn} \\
}

\begin{document}
\maketitle
\begin{abstract}
Perception-enhanced pre-training, particularly through grounding techniques, is widely adopted to enhance the performance of graphical user interface (GUI) agents. However, in resource-constrained scenarios, the format discrepancy between coordinate-oriented grounding and action-oriented reasoning limits the effectiveness of grounding for reasoning tasks. To address this challenge, we propose a query-oriented pivot approach called \textit{query inference}, which serves as a bridge between GUI grounding and reasoning. By inferring potential user queries from a screenshot and its associated element coordinates, query inference improves the understanding of coordinates while aligning more closely with reasoning tasks. Experimental results show that query inference outperforms previous grounding techniques under the same training data scale. Notably, query inference achieves comparable or even better performance to large-scale grounding-enhanced OS-Atlas with less than 0.1\% of training data. Furthermore, we explore the impact of reasoning formats and demonstrate that integrating additional semantic information into the input further boosts reasoning performance. The code is publicly available at \url{https://github.com/ZrW00/GUIPivot}.
\end{abstract}

\section{Introduction}

\begin{figure}[!t]
  \centering
  \includegraphics[width=\linewidth]{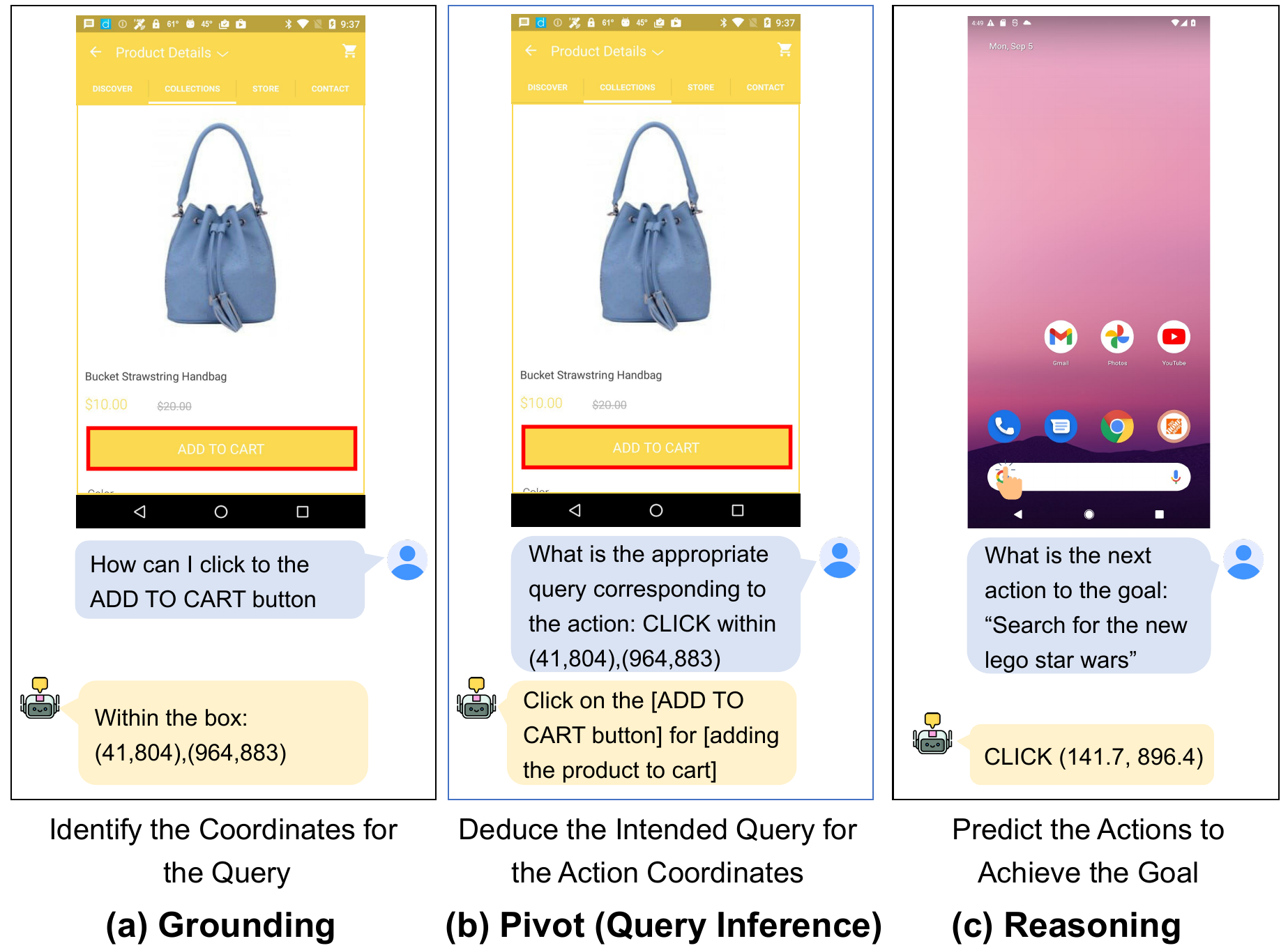}	
  \caption{Illustration of grounding, query inference, and reasoning. Grounding identifies coordinates for the queries, while reasoning predicts the actions to achieve the goal. Query inference deduces the intended user queries for the action coordinates, serving as the pivot approach to smooth grounding and reasoning.\vspace{-0.4cm}}
  \label{fig:intro}
\end{figure}

The development of multimodel large language models (MLLMs)~\citep{yin2023survey,wang2024qwen2,wu2024nextgpt} provides a promising solution for improving the functionality and efficiency of graphical user interface (GUI) agents~\citep{zhang2024you,zhang2024large,ma2024comprehensive}. Since most MLLMs are rarely pre-trained on GUI screenshots, perception-enhanced pre-training tasks on GUI screenshots~\citep{zhang2024ui, you2025ferret, qin2025ui}, particularly through grounding that identifies coordinates for the queries~\citep{wu2025osatlas}, are introduced to improve the understanding of GUI environments. By leveraging continual pre-training on perception-enhanced tasks and supervised fine-tuning (SFT) on reasoning tasks~\citep{rawles2023android, li2024effects, zhang2024android}, MLLMs can serve as the foundation brain of GUI agents, enabling them to navigate within complex GUI environments to predict and execute multiple actions to achieve user-specific goals~\citep{zhang2024large}.

Despite the success of widely-adopted grounding, grounding typically requires large-scale training data~\citep{wu2025osatlas,qin2025ui}. However, in resource-constrained scenarios, such as personalized agents~\citep{cai2024large}, the model scale and available training data are insufficient to support large-scale grounding. Focusing on this resource-constrained scenario where (i) the model scale and (ii) the available training data are constrained for lightweight deployment, we investigate the effectiveness of grounding in such scenarios. 

As we will show later (Section~\ref{sec:pilotInvestigation}), grounding with limited training data leads to minimal improvements in reasoning, highlighting a gap due to the task format discrepancy between coordinate-oriented grounding and action-oriented reasoning in resource-constrained scenarios. This raises a key research question: \textit{Is it possible to bridge the gap between coordinate-oriented grounding and action-oriented reasoning to enhance the performance of GUI agents in resource-constrained scenarios?}

To address the research question, we propose a query-oriented pivot approach, named \textit{query inference}, to serve as a bridge between GUI grounding and reasoning. As shown in Figure~\ref{fig:intro}, query inference deduces the intended user queries corresponding to action coordinates, enhancing the understanding of coordinates and GUI layouts while better aligning with action-oriented reasoning tasks. This task format resembles the reverse process of grounding, enabling easier construction of query inference data by refining existing grounding data.

Experimental results demonstrate that query inference outperforms grounding with the same data scale. 
Additionally, when employed as a pivot between grounding and reasoning, query inference further enhances action prediction. Notably, query inference can achieve comparable or even better performance to the large-scale grounding-enhanced OS-Atlas~\citep{wu2025osatlas} with less than 0.1\% of training data. Furthermore, we explore the efficiency of query inference in conjunction with chain-of-thought (CoT) enhanced reasoning \citep{zhang2024android, sun2024genesis}, revealing that incorporating additional semantic perception into inputs further improves reasoning. 

Our contributions are summarized as follows:

(i) We investigate the effectiveness of grounding in resource-constrained scenarios and find its minimal improvements on reasoning, revealing a significant gap due to the task format discrepancy between coordinate-oriented grounding and action-oriented reasoning (Section~\ref{sec:pilotInvestigation}).

(ii) To bridge this gap, we propose a query-oriented pivot approach, named \textit{query inference}, to smooth grounding and reasoning. Query inference deduces the intended queries corresponding to action coordinates, thereby enhancing the understanding of coordinates while aligning better with action-oriented reasoning (Section~\ref{sec:methodology}). 

(iii) Through extensive experiments, we validate the effectiveness and potential of query inference in resource-constrained scenarios. Notably, query inference achieves performance comparable to large-scale grounding-enhanced OS-Atlas with less than 0.1\% of the training data (Section~\ref{sec:experiments}).

\section{Related Works}\label{sec:relatedWorks}
In this section, we review related works that form the basis of this work from three perspectives: MLLM-powered GUI agents, perception-enhanced pre-training, and CoT enhanced reasoning.

\paragraph{\textbf{MLLM-powered GUI Agents.} }
The advent of MLLMs~\citep{yin2023survey, chen2024internvl, wang2024qwen2} has flourished promising opportunities to develop GUI-based agents~\citep{cheng2024seeclick, hong2024cogagent, gou2025navigating}. Unlike traditional text-based perception, which typically require system-level permissions to access textual representations of GUI environments~\citep{zhou2024webarena, deng2024mind2web}, MLLM-powered GUI agents directly utilize the vision modules to perceive and interact directly within GUI environments through human-like actions, such as \texttt{CLICK}, \texttt{TYPE}, and \texttt{SCROLL}, without relying on programmatic interactions~\citep{sun2024survey} or API calls~\citep{wu2024oscopilot, zhang2024ufo}.

\paragraph{\textbf{Perception-enhanced Pre-training.} }
Since most open-source MLLMs are primarily pre-trained on natural images and struggle to perceive high-density GUI environments~\citep{wu2025osatlas}, perception-enhanced pre-training is widely adopted to improve GUI understanding. One of the most prevalent pre-training tasks is grounding~\cite{wu2025osatlas, qian2024chatdev}, which identifies and localizes GUI elements corresponding to user queries. Other tasks include GUI referring~\cite{zhang2024ui, you2025ferret}, which generates descriptions for specific GUI elements, and screen question answering~\cite{baechler2024screenai, chen2024webvln}, which answers questions about screen contents and functionalities. However, perception-enhanced pre-training typically requires large-scale training data, and its feasibility in resource-constrained scenarios remains underexplored. \looseness=-1

\paragraph{\textbf{CoT Enhanced Reasoning.} }
Recently, CoT~\citep{wei2022chain, zhang2024multimodal, chu2024navigate} is introduced to GUI agents to enhance reasoning~\citep{zhang2024android, sun2024genesis}. By leveraging proprietary MLLMs as annotation models~\citep{achiam2023gpt, bai2023qwen}, semantic information is automatically generated to enrich training data for improved reasoning. Specifically, explanations for GUI environments, such as screen descriptions (SD), previous action results (PAR), and GUI layouts~\citep{ma2024comprehensive} are incorporated into inputs to enhance perception, while intermediate reasoning results like action thoughts (AT) and next action descriptions (AD) are introduced into outputs to improve reasoning process.

\section{Preliminary Study}\label{sec:pilotInvestigation}
In this section, we describe the formulation of grounding and reasoning in Section~\ref{subsec:formula} and investigate the effectiveness of grounding with limited data for reasoning in Section~\ref{subsec:gap}.

\subsection{Formulation of Grounding and Reasoning}\label{subsec:formula}
Grounding, a widely adopted perception-enhanced pre-training task, aims to localize the coordinates $c$ of specific GUI elements based on the perception of screenshots $s$ and low-level unintended queries $q$. 
Specifically, $q$ can consist of explicit instructions, such as \textit{``click the clock icon''}, which directly refer to identifiable elements, or more complex, implicit instructions that require additional reasoning,  like \textit{``click on the home button at top left''}~\citep{bai2021uibert}, which necessitate understanding of both the query context and the relative positioning of the elements within the interface. The coordinates $c$ can be represented as either points or bounding boxes. Formally, grounding can be represented as:
\begin{equation}
  \label{equ:grounding}
    \mathcal{G}: \{\langle s, q \rangle\} \to \{c\}.
\end{equation}

Based on the perception of GUI environments, reasoning predicts a chain of actions to achieve the high-level final goals. At step $i$, the agent perceives the current screenshot $s_i$ along with historical actions $\{a_{<i}\}$ to predict current action $a_i$ to achieve the final goal $g$. During reasoning, $a_i$ typically consists of action type $t$, and action parameters $p$, which may include typed text or coordinates $c$~\citep{wu2025osatlas}. Recently, optional CoT components like intermediate reasoning thoughts $r$ are also introduced into $a_i$ to enhance reasoning. Therefore, reasoning at step $i$ can be formulated as:
\begin{equation}
  \label{equ:reasoning}
    \mathcal{R}: \left\{\langle s_i, \left\{a_{<i}\right\}, g \rangle \right\} \to \left\{a_i \right\}.
\end{equation}

As illustrated in Equation~\ref{equ:reasoning}, reasoning is action-oriented and requires profound comprehension of high-level user intent, whereas grounding is coordinate-oriented and only aligns low-level queries with coordinates within a single screenshot, lacking perception of high-level intent. This format discrepancy creates a gap between grounding and reasoning. While large-scale training data can help mitigate this gap, it may be particularly pronounced in resource-constrained scenarios.

\subsection{Grounding with Small Scale Data}\label{subsec:gap}

\begin{table}
  \centering
  \small
  \setlength{\tabcolsep}{1.5pt}
  \resizebox{\linewidth}{!}{ 
  \begin{tabular}{ccccccc}
  \toprule
  \multirow{2}{*}{Pipeline} & \multicolumn{2}{c}{AndroidControl-L} & \multicolumn{2}{c}{AndroidControl-H} & \multicolumn{2}{c}{AITZ} \\
                            & TMR$\uparrow$              & AMR$\uparrow$             & TMR$\uparrow$               & AMR$\uparrow$             & TMR$\uparrow$       & AMR$\uparrow$      \\ \midrule
  SFT                       & 96.84             & \textbf{84.33}              & 80.38              & 65.23              & 75.76      & 61.43       \\
  Grounding+SFT             & \textbf{96.85}             & 83.88              & \textbf{81.37}              & \textbf{65.57}              & \textbf{81.58}      & \textbf{63.48}       \\ \midrule
  Atlas-7B+SFT               & 94.96             & 86.80              & 81.78              & 68.65              & 82.03      & 67.04       \\ \bottomrule
  \end{tabular}
  }
  \caption{Performance on mobile agent benchmarks with and without grounding on UIBERT. AndroidControl-L refers to the scenario where both low-level step instructions and high-level goals are provided as inputs, while AndroidControl-H indicates that only high-level goals are provided. The optimal values are \textbf{bolded}.\vspace*{-0.4cm}}
  \label{tab:groundingGap}
\end{table}

While extensive studies demonstrate the effectiveness of grounding in enhancing reasoning with large-scale grounding data (typically exceeding 10 million)~\citep{wu2025osatlas,qin2025ui}, grounding with limited data in resource-constrained scenarios, such as personalized mobile agents, remains underexplored. As illustrated in Section~\ref{subsec:formula}, grounding provides perception for low-level queries but leaves a gap to action-oriented reasoning. To demonstrate this, we evaluate the reasoning performance with and without grounding on limited grounding data.

\begin{figure*}[!htb]
  \centering
  \includegraphics[width=\linewidth]{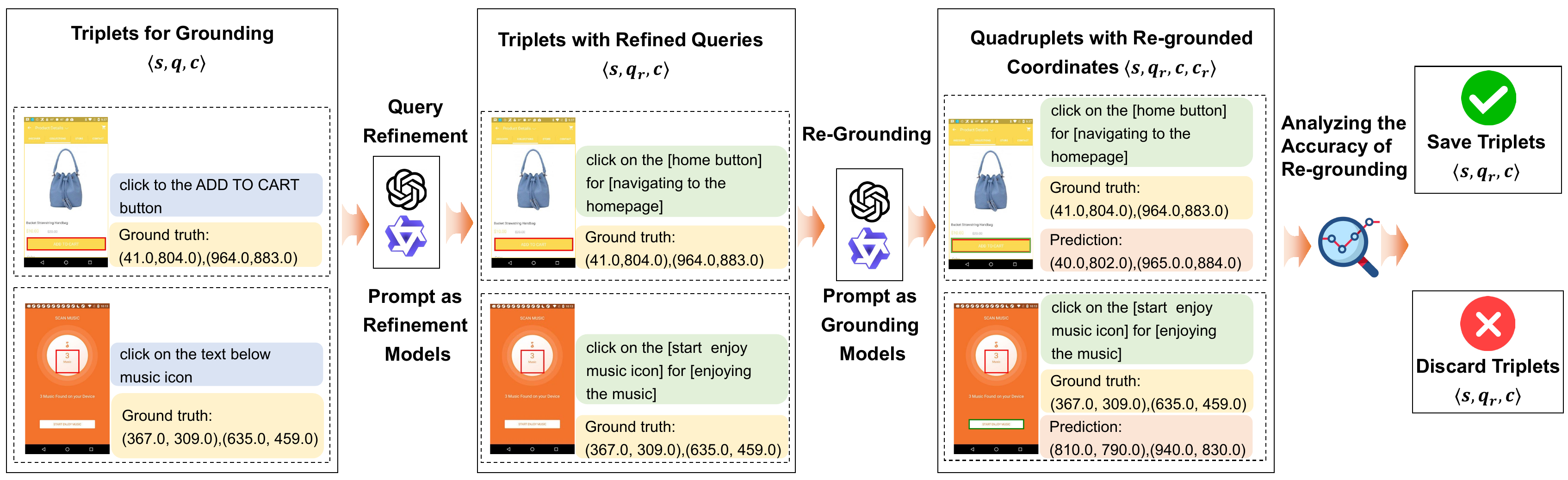}	
  \caption{Three-step Pipeline for constructing samples for query inference. First, we utilize proprietary MLLMs to refine low-level unintended queries into intented formated queries based on corresponding coordinates and screenshots. Second, We utilize proprietary MLLMs for re-grounding based on the refined queries. Finally, we analyze the accuracy of the predicted coordinates to decide whether to save the sample. \vspace*{-0.6cm}}
  \label{fig:pipeline}
\end{figure*}

Specifically, we select UIBERT~\citep{bai2021uibert}, which contains about 10,000 instances of grounding data, as the grounding dataset for resource-constrained scenarios. UIBERT is a subset of the OS-Atlas~\citep{wu2025osatlas} grounding dataset with more than 13 million samples. 
Following \citet{wu2025osatlas} and \citet{qin2025ui}, we choose the widely adopted Qwen2-VL-7B-Instruct~\citep{wang2024qwen2} as the foundation MLLM for grounding. After obtaining the grounding-enhanced model, we fine-tune it on two mobile agent benchmarks, AndroidControl~\citep{li2024effects} and AITZ~\citep{zhang2024android}.  Specifically, we evaluate AndroidControl in two settings: with both low-level instructions and high-level goals (denoted as AndroidControl-L), and with only high-level goals (denoted as AndroidControl-H). Then, we evaluate action prediction performance with and without grounding by utilizing action type match rate (TMR) and exact action match rate (AMR). For comparasion, we also fine-tune OS-Atlas-Base-7B (dubbed as Atlas)~\citep{wu2025osatlas} on these benchmarks to access its action prediction performance.

The action prediction results are presented in Table~\ref{tab:groundingGap}. We find that grounding with limited data leads to minimal improvement.
Specifically, on AndroidControl without low-level step instructions, grounding on UIBERT only leads to a negligible 0.34\% improvement on AMR. Conversely, on AndroidControl with low-level step instructions, grounding even results in negative optimization. Significant improvements in AMR are only observed when large-scale grounding data are used. These results highlight the gap between coordinate-oriented grounding and action-oriented reasoning in resource-constrained scenarios, underscoring the demand to bridge this gap. 

\section{Methodology}\label{sec:methodology}
Findings in Section~\ref{sec:pilotInvestigation} indicate that the task format discrepancy between coordinate-oriented grounding and action-oriented reasoning leads to the minimal improvements of grounding in resource-constrained scenarios. To address the challenge, we propose \textit{query inference}, a query-oriented task to smooth grounding and reasoning.

As illustrated in Section~\ref{subsec:formula}, reasoning requires profound comprehension of user intented query. Intuitively, a query-oriented task that deduces user queries from corresponding action coordinates may effectively enhance query comprehension. This can be simply implemented by reversing the grounding process. However, existing grounding queries are typically unintended, making it challenging to align them with high-level reasoning instructions.

Inspired by recent works that leverages proprietary MLLMs as the annotation models to construct CoT annotations~\citep{zhang2024android} and instantiate task trajectory data~\cite{sun2024genesis}, we utilize proprietary MLLMs as refinement models to transform low-level unintended queries into intended, properly formatted queries. Subsequently, we employ MLLMs as grounding models to filter high-quality refined queries. Consequently, we propose a three-step pipeline: query refinement, re-grounding, and analyzing the accuracy of re-grounding, to construct samples for query inference, as shown in Figure~\ref{fig:pipeline}.

\paragraph{\textbf{Query Refinement.}}
First, we utilize the proprietary MLLM, Qwen-VL-Max~\citep{bai2023qwen}, as the refinement model $\mathcal{M}_{r}$. We prompt $\mathcal{M}_{r}$ to transform the low-level unintended queries $q$ into intended queries $q_{r}$ in the format: \textit{click on the} \texttt{[element\_name]} \textit{for} \texttt{[purpose]}, based on corresponding coordinates $c$ and screenshots $s$ from the grounding data. The refinement process aims to deduce the intention behind actions interacting with the coordinate-specified elements. Formally, the refinement process can be represented as:
\begin{equation}
  \label{equ:refinement}
    \mathcal{M}_{r}: \{\langle s, q, c \rangle\} \to \{q_r\}.
\end{equation}

\paragraph{\textbf{Re-grounding.}}
Automated refinement may introduce incorrect information. Therefore, inspecting the refined data is crucial to ensure data quality. Specifically, we utilize Qwen-VL-Max as the grounding model $\mathcal{M}_{g}$, prompting $\mathcal{M}_{g}$ to localize the coordinates $c_r$ for further analysis based on the refined queries $q_r$ and the corresponding screenshots $s$. The process is formulated as: 
\begin{equation}
  \label{equ:regrounding}
    \mathcal{M}_{g}: \{\langle s, q_r \rangle\} \to \{c_r\}.
\end{equation}

\paragraph{\textbf{Analyzing the Accuracy of Re-grounding.}}
After obtaining $c_r$, we analyze its accuracy compared to the ground-truth coordinates $c$ to filter out incorrect re-grounding samples corresponding to low-quality refined queries. Similar to the grounding evaluation, we establish an indicator $\mathcal{I}$ to determine whether the center point of $c_r$ lies within the bounding box represented by $c$, as illustrated in Equation~\ref{equ:analysisRegrounding}. If so, the triplet $\langle s, q_r, c \rangle$ is retained as a data sample for query inference; otherwise, the sample is discarded. Finally, the dataset consists of triplets $\langle s, q_r, c \rangle$ for query inference is obtained.
\begin{equation}
  \label{equ:analysisRegrounding}
  \mathcal{I}(c_r, c) = 
  \begin{cases} 
  1, & \text{if the center of } c_r \text{ is inside } c, \\
  0, & \text{otherwise}.
  \end{cases}
\end{equation}

Subsequently, we utilize the dataset to train the foundation MLLM on query inference task prior to reasoning SFT, as shown in Equation~\ref{equ:querysummary}, enhancing the comprehension of user intention to align with reasoning while maintaining sensitivity to the coordinates. Finally, the gap between grounding and reasoning is bridged by query inference.
\begin{equation}
  \label{equ:querysummary}
    \mathcal{Q}: \{\langle s, c \rangle\} \to \{q_r\}.
\end{equation}

\section{Experiments}\label{sec:experiments}
This section evaluates the effectiveness of query inference. We first outline the experimental setup in Section~\ref{subsec:expsetup}. 
Subsequently, in Section~\ref{subsec:mainResults}, we present the empirical results. Finally, in Section~\ref{subsec:analysis}, we analyze the experimental findings.

\subsection{Experimental Setup}\label{subsec:expsetup}
\paragraph{\textbf{Datasets.}}
In alignment with Section~\ref{subsec:gap}, for perception-enhanced pre-training, we select UIBERT~\citep{bai2021uibert} as the dataset for grounding and constructing the query inference dataset. The final query inference dataset, refined from UIBERT, consists of 9,570 triplets of $\langle s, q_r, c \rangle$, with examples provided in Appendix~\ref{subappendix:refinedUIBERT}. For fairness, we extract corresponding samples from the original UIBERT dataset as grounding training data. For reasoning, we choose two public mobile agent benchmarks: AndroidControl~\citep{li2024effects} and AITZ~\cite{zhang2024android}. We utilize the training subset of the benchmarks to SFT and the test subset for evaluation. Dataset details AndroidControl and AITZ benchmarks are provided in Appendix~\ref{subappendix:datasets}.

\paragraph{\textbf{Models.}}
In alignment with Section~\ref{subsec:gap}, we adopt Qwen2-VL-7B-Instruct (dubbed as Qwen)~\cite{wang2024qwen2} as the foundation MLLM for grounding and query inference and subsequent reasoning SFT. Additionally, to compare the action prediction performance of large-scale grounding-enhanced models, we also fine-tune OS-Atlas-Base-7B (dubbed as Atlas)~\citep{wu2025osatlas}, which is trained on over 13 million grounding samples, on mobile agent benchmarks for comparison.

\paragraph{\textbf{Metrics.}} 
We evaluate final action prediction accuracy to assess the impact of grounding and query inference on reasoning performance. Specifically, in alignment with Section~\ref{subsec:gap}, we evaluate action prediction accuracy by adopting two commonly used metrics for GUI agents that assess the accuracy of action type match rate (TMR) and exact action match rate (AMR). TMR measures the match rate between predicted action types (e.g., \texttt{PRESS}, \texttt{SCROLL}) and ground truth types. AMR evaluates whether the predicted action exactly matches the ground truth within a single step, considering both action type $t$ and optional parameters $p$ (e.g., coordinates, app names, and text input). An action is considered an exact match only when $t$ and $p$ align perfectly with the ground truth. Details on AMR evaluation are provided in Appendix~\ref{subappendix:amr}. 

\begin{table*}
  \centering
  \small
  \setlength{\tabcolsep}{2.2pt}
  \begin{tabular}{@{}ccccccccccccc@{}}
      \toprule
      \multirow{2}{*}{Dataset}          & \multirow{2}{*}{\makecell{Foundation \\ Model}} & \multirow{2}{*}{Approach} & \texttt{SCROLL} & \multicolumn{2}{c}{\texttt{CLICK}} & \multicolumn{2}{c}{\texttt{TYPE}} & \texttt{PRESS} & \multicolumn{2}{c}{\texttt{OPENAPP}} & \multicolumn{2}{c}{TOTAL} \\ \cmidrule(l){4-13} 
                                        &                                   &                               & TMR$\uparrow$    & TMR$\uparrow$         & AMR$\uparrow$         & TMR$\uparrow$         & AMR$\uparrow$        & TMR$\uparrow$   & TMR$\uparrow$          & AMR$\uparrow$          & TMR$\uparrow$         & AMR$\uparrow$         \\ \midrule
      \multirow{5}{*}{AndroidControl-L} & \multirow{4}{*}{Qwen}             & /                             & \textbf{91.49}  & 97.26       & 75.07       & \textbf{98.55}       & \textbf{88.95}      & \textbf{97.96} & \textbf{99.84}        & 83.55        & \underline{96.84}       & 84.33       \\
                                        &                                   & $\mathcal{G}$                 & \underline{91.25}  & \textbf{97.42}       & 76.01       & 96.99       & 77.69      & \underline{97.67} & 99.34        & 85.86        & \textbf{96.85}       & 83.88       \\
                                        &                                   & $\mathcal{Q}$                 & 91.08  & \underline{97.32}       & \textbf{78.95}       & \underline{97.78}       & 79.59      & \underline{97.67} & 99.51        & \underline{86.02}        & 96.79       & \underline{85.45}       \\
                                        &                                   & $\mathcal{G}+\mathcal{Q}$     & 91.08  & 96.49       & \underline{78.87}       & 97.31       & \underline{79.91}      & 97.08 & \underline{99.67}        & \textbf{88.16}        & 96.48       & \textbf{85.70}       \\ \cmidrule(l){2-13} 
                                        & Atlas                             & /                             & 91.58  & 97.48       & 85.69       & 97.38       & 79.59      & 97.67 & 99.84        & 83.39        & 94.96       & 86.80       \\ \midrule
      \multirow{5}{*}{AndroidControl-H} & \multirow{4}{*}{Qwen}             & /                             & \textbf{60.94}  & 85.26       & 59.83       & 87.82       & \textbf{69.92}      & 56.27 & 90.13        & 75.66        & 80.38       & 65.23       \\
                                        &                                   & $\mathcal{G}$                 & \underline{59.95}  & 85.87       & 61.17       & \textbf{90.51}       & 55.22      & \textbf{61.52} & \textbf{92.76}        & 75.99        & 81.37       & 65.57       \\
                                        &                                   & $\mathcal{Q}$                 & 57.64  & \underline{87.31}       & \underline{63.11}       & 71.77       & 54.11      & \underline{58.69} & \underline{91.78}        & \textbf{77.14}        & \textbf{81.68}       & \underline{66.11}       \\
                                        &                                   & $\mathcal{G}+\mathcal{Q}$     & 58.79  & \textbf{87.76}       & \textbf{63.83}       & 89.72       & 53.32      & 57.14 & 90.95        & \underline{76.48}        & \underline{81.59}       & \textbf{66.24}       \\ \cmidrule(l){2-13} 
                                        & Atlas                             & /                             & 61.85  & 85.28       & 65.43       & 91.77       & 55.70      & 67.93 & 94.74        & 82.24        & 81.78       & 68.65       \\ \midrule
      \multirow{5}{*}{AITZ}             & \multirow{4}{*}{Qwen}             & /                             & 59.73  & 81.40       & 63.23       & 86.40       & \textbf{50.40}      & 71.32 & /            & /            & 75.76       & 61.43       \\
                                        &                                   & $\mathcal{G}$                 & \underline{60.39}  & 86.51       & 66.88       & 86.80       & 48.60      & 73.58 & /            & /            & 81.58       & 63.48       \\
                                        &                                   & $\mathcal{Q}$                 & 60.23  & \textbf{87.57}       & \textbf{67.80}        & \textbf{88.20}       & 48.60      & \underline{77.36} & /            & /            & \underline{82.26}       & \underline{66.62}       \\
                                        &                                   & $\mathcal{G}+\mathcal{Q}$     & \textbf{63.06}  & \underline{87.54}       & \underline{67.65}       & \underline{87.80}       & \underline{48.80}      & \textbf{78.49} & /            & /            & \textbf{82.54}       & \textbf{66.91}       \\ \cmidrule(l){2-13} 
                                        & Atlas                             & /                             & 65.39  & 86.37       & 67.54       & 88.40        & 49.80       & 76.60  & /            & /            & 82.03       & 67.04       \\ \bottomrule
  \end{tabular}
  \caption{Overall and type-wise action prediction performance when trained with grounding, query inference as the alternative task, and query inference as the pivot task on AndroidControl and AITZ. The optimal and the suboptimal results are \textbf{bolded} and \underline{underlined}, respectively.\vspace{-0.6cm}}
  \label{tab:reasoningImprovements}
\end{table*}

\paragraph{\textbf{Implementation Details.}}
Following~\citet{wu2025osatlas}, we normalize all coordinates to the range [0, 1000]. For reasoning SFT, following~\citet{wu2025osatlas}, we unify the action space into three basic actions: \texttt{CLICK}, \texttt{TYPE}, and \texttt{SCROLL}, along with custom actions like \texttt{OPENAPP} for AndroidControl and AITZ.  We adopt LLaMa-Factory~\citep{zheng2024llamafactory} framework to train on grounding and query inference, as well as SFT on mobile agent benchmarks. The learning rate is uniformly set to $1\times 10 ^{-5}$, with training epochs set to 5 for grounding and query inference and 3 for SFT on reasoning, respectively. During testing, we adopt flash-attn~\citep{dao2023flashattention2} for acceleration. Detailed prompts for query refinement, grounding, query inference, and action prediction are provided in Appendix~\ref{appendix:prompts}.

\subsection{Main Results}\label{subsec:mainResults}
Table~\ref{tab:reasoningImprovements} presents the main results on overall and type-wise action prediction performance.

Specifically, we apply the foundation models in four settings: (i) skip perception-enhanced pre-training, where the model is directly fine-tuned on the mobile agent benchmarks; (ii) grounding, denoted as $\mathcal{G}$, which is trained for grounding on UIBERT, followed by subsequent reasoning SFT; (iii) query inference as the alternative task, denoted as $\mathcal{Q}$, which is trained for query inference on the refined UIBERT dataset and followed by subsequent reasoning SFT; (iv) query inference as the pivot task, denoted as $\mathcal{G} + \mathcal{Q}$, where the model is trained on half of the refined UIBERT dataset for grounding and the other half for query inference, followed by subsequent reasoning SFT. To compare the action prediction performance of large-scale grounding-enhanced models, we also fine-tune Atlas on mobile agent benchmarks and evaluate its action prediction performance.

Our key findings are as follows:

(i) Query inference outperforms grounding with the same data scale. While grounding yields the optimal TMR on AndroidControl with low-level step instructions, the improvement over other settings is minimal. Conversely, adopting query inference as either alternative task or pivot task yields over 1\% improvements to directly SFT on AndroidControl, outperforming grounding. While on AITZ, the improvements are more substantial, exceeding 5\%. These findings highlight the effectiveness of query inference in resource-constrained scenarios.

(ii) Adopting query inference as the pivot task further improves reasoning. Generally, adopting query inference as the pivot task achieves the optimal AMRs across four settings of Qwen model, surpassing its use as the alternative task. These indicate that adopt query inference as pivot task smooths grounding and reasoning, enhancing the understanding of both coordinates and user queries, thereby improving reasoning performance.

(iii) Adopting query inference as the pivot task achieves performance comparable to the large-scale grounding-enhanced Atlas. Specifically, adopting query inference as the pivot task yields comparable AMRs to Atlas on AndroidControl, with a minimal discrepancy (around 0.1\%) on AITZ. Furthermore, the TMR of adopting query inference as the pivot task on AndroidControl with low-level step instructions and AITZ even surpasses that of directly fine-tuning Atlas. These suggest that query inference can achieve comparable performance to large-scale grounding with less than 0.1\% of training data, indicating it as a more effective approach in resource-constrained scenarios.

(iv) Query inference most significantly improves performance in the critical \texttt{CLICK} actions, consistently yielding either optimal or suboptimal results when adopted as the alternative or pivot task. For other action types, query inference demonstrates superior or comparable  performance. However, for \texttt{TYPE} actions, including Atlas, AMR experiences significant degradation compared to directly fine-tuning Qwen on mobile agent benchmarks. This may be attributed to the vertical tuning on GUI scenarios, which could hinder the instruction-following capability of the model. Despite this, query inference generally improves action prediction performance across most action types.

\subsection{Analysis}\label{subsec:analysis}
In this section, we present further discussions and analysis to the detailed experiment results. We investigate the impact of training data scale on overall action prediction performance in Section~\ref{subsubsec:datascale}. Additionally, we evaluate the improvements of query inference when combined with CoT-enhanced reasoning in Section~\ref{subsubsec:CoTEnhancedReasoning}.

\begin{figure}
  \centering
  \includegraphics[width=\linewidth]{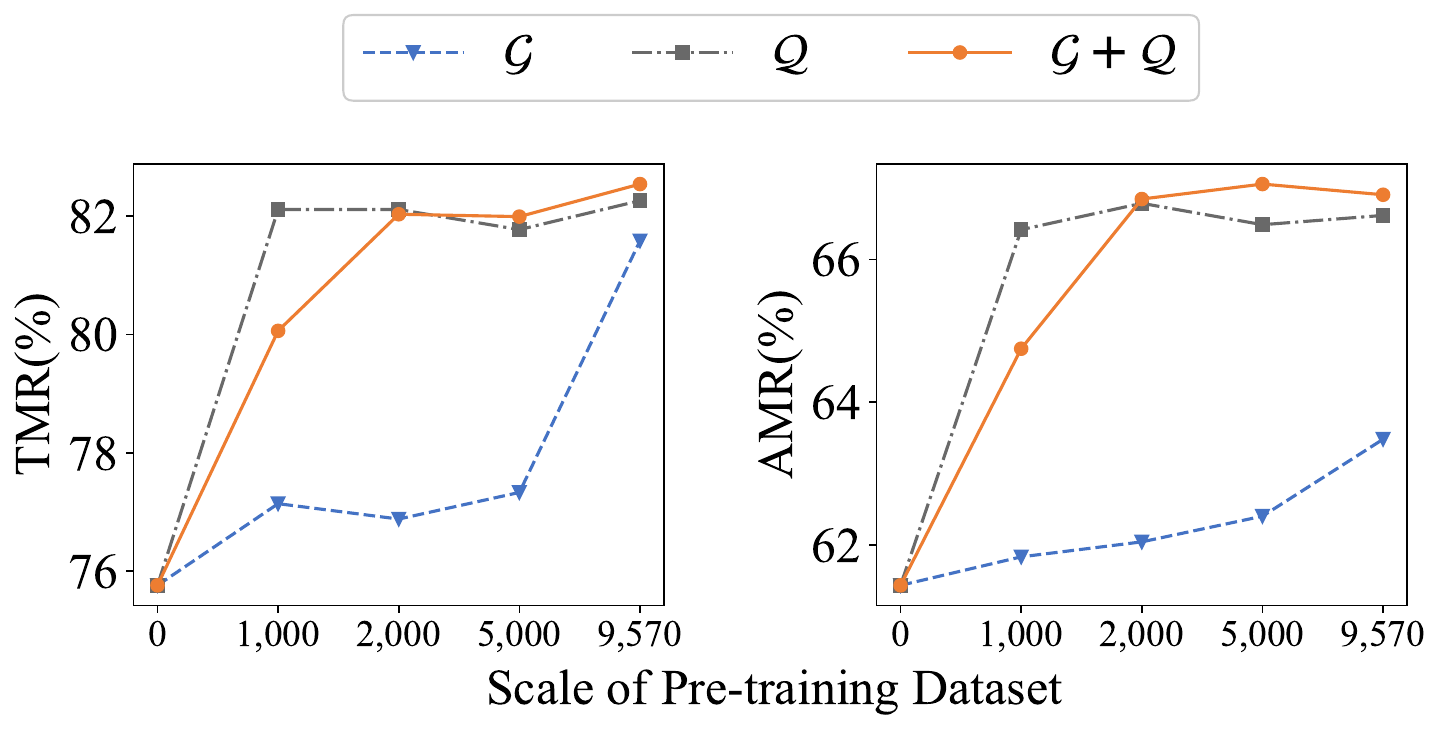}
  \caption{The overall action prediction performance on AITZ when trained with grounding, query inference as the alternative task, and query inference as the pivot task across various data scales.}
  \label{fig:dataScaling}
\end{figure}

\subsubsection{Influence of Training Data Scale}\label{subsubsec:datascale}
To thoroughly investigate the effectiveness of query inference under various data scales in resource-constrained scenarios, we randomly extract 1,000, 2,000, and 5,000 samples from the original refined query inference dataset for training, followed by subsequent fine-tuning on AITZ. This enables us to investigate the overall action prediction performance across these varying training data scales. The results are shown in Figure~\ref{fig:dataScaling}. From these results, we draw the following conclusions:

(i) Query inference is generally more effective than grounding in resource-constrained scenarios. For grounding, the performance of action prediction increases gradually as the data scale expands, demonstrating a steady but slower improvement with the availability of more samples. In contrast, query inference exhibits a much faster rate of performance improvement, reaching its peak performance with approximately 2,000 training samples. This highlights the efficiency of query inference with limited data, consistently outperforming grounding across all tested data scales.

(ii) Query inference as the pivot task performs better with larger datasets. When more than 5,000 training samples are utilized, query inference as the pivot task yields better performance. However, with smaller datasets, query inference as the alternative task performs better.

(iii) Grounding is more sensitive to data scale. A significant performance increase is observed with grounding when training exceeds 5,000 samples, indicating that grounding benefits substantially from large-scale training, consistent with the proven success of grounding in such scenarios~\citep{wu2025osatlas,qin2025ui}.

\subsubsection{Combination with CoT-enhanced Reasoning}\label{subsubsec:CoTEnhancedReasoning}

\begin{table*}[!t]
  \centering
  \small
  \setlength{\tabcolsep}{5.5pt}
    \begin{tabular}{@{}cccccccccccccc@{}}
      \toprule
      \multirow{2}{*}{Pre-training} & \multirow{2}{*}{ID}               & \multicolumn{2}{c}{Input} & \multicolumn{2}{c}{Output} & \texttt{SCROLL} & \multicolumn{2}{c}{\texttt{CLICK}} & \multicolumn{2}{c}{\texttt{TYPE}} & \texttt{PRESS} & \multicolumn{2}{c}{TOTAL} \\ \cmidrule(l){3-14} 
                                                &  & SD          & PAR         & AT           & AD          & TMR$\uparrow$    & TMR$\uparrow$         & AMR$\uparrow$         & TMR$\uparrow$         & AMR$\uparrow$        & TMR$\uparrow$   & TMR$\uparrow$         & AMR$\uparrow$         \\ \midrule
      \multirow{10}{*}{$\mathcal{G}$}    &   1      &             &             &              &             & 60.39  & 86.51       & \underline{66.88}       & \underline{86.80}       & 48.60      & 73.58 & 81.58       & 63.48       \\ \cmidrule(l){2-14}
                                                  & 2 &$\checkmark$           &             &              &             & 60.40  & 85.96       & 66.56       & \textbf{88.80}       & \underline{49.00}      & \underline{73.96} & 81.22       & 65.77       \\
                                                  &  3 &           & $\checkmark$           &              &             & \textbf{60.73}  & \underline{86.95}       & \textbf{67.32}       & 86.40       & 47.20      & \textbf{75.47} & \underline{81.75}       & \textbf{66.23}       \\
                                                   & 4 & $\checkmark$           & $\checkmark$           &              &             & 60.23  & 85.64       & 66.04       & \textbf{88.80}       & \textbf{51.00}      & 73.58 & 81.14       & \underline{65.79}       \\ \cmidrule(l){2-14}
                                                  &  5 &           &             & $\checkmark$            &             & 53.24  & 84.28       & 61.51       & 83.80       & 48.00      & 72.08 & 77.10       & 60.12       \\
                                                  & 6 &            &             &              & $\checkmark$           & \underline{60.57}  & \textbf{88.67}       & 65.83       & 85.20       & 48.00      & \underline{73.96} & \textbf{82.36}       & 65.09       \\
                                                  &  7 &           &             & $\checkmark$            & $\checkmark$           & 50.75  & 72.33       & 52.12       & 80.60       & 45.00      & 69.81 & 69.60       & 54.13       \\ \cmidrule(l){2-14}
                                                  & 8 &$\checkmark$           &             & $\checkmark$            & $\checkmark$           & 50.42  & 73.61       & 53.76       & 82.00       & 44.40      & 69.81 & 70.07       & 54.59       \\
                                                  &  9 &           & $\checkmark$           & $\checkmark$            & $\checkmark$           & 50.92  & 72.40       & 52.56       & 82.40       & 46.40      & 70.57 & 70.00       & 54.59       \\
                                                  & 10 &$\checkmark$           & $\checkmark$           & $\checkmark$            & $\checkmark$           & 50.58  & 73.90       & 54.09       & 84.00       & 45.20      & 69.81 & 70.17       & 54.59       \\ \midrule
      \multirow{10}{*}{$\mathcal{Q}$}  & 1           &             &             &              &             & 60.23  & 87.57       & 67.80       & 88.20       & 48.60      & \textbf{77.36} & 82.26       & 66.62       \\ \cmidrule(l){2-14}
                                                  & 2  &$\checkmark$           &             &              &             & \underline{61.73}  & 87.61       & \textbf{68.46}       & 88.80       & \underline{49.40}      & \underline{76.98} & \underline{82.77}       & 66.62       \\
                                                  &  3 &           & $\checkmark$           &              &             & 61.23  & \underline{87.76}       & \underline{67.84}       & \underline{89.60}       & 49.20      & \underline{76.98} & \textbf{82.87}       & \textbf{67.06}       \\
                                                  & 4 &$\checkmark$           & $\checkmark$           &              &             & \textbf{63.89}  & 85.78       & 66.89       & \textbf{90.60}       & \textbf{50.20}      & \textbf{77.36} & 82.13       & \underline{66.91}       \\ \cmidrule(l){2-14}
                                                  &  5 &            &             & $\checkmark$            &             & 50.25  & 84.61       & 63.71       & 84.20       & 47.40      & 72.08 & 77.05       & 61.05       \\
                                                  &  6 &            &             &              & $\checkmark$           & 58.74  & \textbf{89.00}       & 66.52       & 86.40       & 47.00      & 74.72 & 82.15       & 64.97       \\
                                                  &  7 &           &             & $\checkmark$            & $\checkmark$           & 49.42  & 73.65       & 53.11       & 81.60       & 45.40      & 72.83 & 70.58       & 54.85       \\ \cmidrule(l){2-14}
                                                  & 8 & $\checkmark$           &             & $\checkmark$            & $\checkmark$           & 52.75  & 72.77       & 52.92       & 82.60       & 46.80      & 70.19 & 70.03       & 54.74       \\
                                                  &  9 &           & $\checkmark$           & $\checkmark$            & $\checkmark$           & 51.41  & 73.21       & 53.33       & 82.40       & 46.40      & 73.21 & 70.53       & 55.21       \\
                                                  &  10 &  $\checkmark$           & $\checkmark$           & $\checkmark$            & $\checkmark$           & 50.42  & 72.84       & 52.81       & 81.80       & 43.80      & 69.43 & 69.71       & 54.09       \\ \midrule
      \multirow{10}{*}{$\mathcal{G}+\mathcal{Q}$} & 1 &            &             &              &             & \textbf{63.06}  & 87.54       & 67.65       & 87.80       & 48.80      & \textbf{78.49} & \underline{82.54}       & \underline{66.91}       \\ \cmidrule(l){2-14} & 2
                                                  & $\checkmark$           &             &              &             & \underline{61.73}  & \underline{87.61}       & \textbf{67.98}       & \underline{89.00}       & \underline{50.20}      & 75.85 & \textbf{87.77}       & \textbf{67.27}       \\
                                                  & 3 &            & $\checkmark$           &              &             & 60.73  & 87.43       & 67.25       & \textbf{89.80}       & \textbf{51.60}      & 75.85 & 82.35       & 66.62       \\
                                                  & 4 &$\checkmark$           & $\checkmark$           &              &             & 61.23  & 87.06       & \underline{67.95}       & 88.60       & 47.60      & \underline{76.23} & 80.88       & 65.47       \\ \cmidrule(l){2-14} 
                                                  &   5 &          &             & $\checkmark$            &             & 52.25  & 84.14       & 62.83       & 81.60       & 47.40      & 72.83 & 77.34       & 61.39       \\
                                                  & 6 &            &             &              & $\checkmark$           & 60.40  & \textbf{88.78}       & 65.57       & 86.60       & 49.20      & 71.70 & 82.26       & 64.86       \\
                                                  & 7 &            &             & $\checkmark$            & $\checkmark$           & 52.91  & 72.04       & 52.12       & 82.20       & 48.20      & 75.47 & 70.17       & 55.03       \\ \cmidrule(l){2-14}
                                                  & 8 &$\checkmark$           &             & $\checkmark$            & $\checkmark$           & 50.25  & 72.62       & 52.81       & 81.80       & 46.60      & 70.57 & 69.39       & 54.19       \\
                                                  & 9 &            & $\checkmark$           & $\checkmark$            & $\checkmark$           & 50.42  & 72.95       & 54.02       & 83.80       & 49.00      & 71.70 & 70.41       & 55.75       \\
                                                  & 10 &$\checkmark$           & $\checkmark$           & $\checkmark$            & $\checkmark$           & 51.58  & 73.83       & 53.03       & 82.00       & 46.00      & 70.94 & 70.62       & 54.76       \\ \bottomrule
      \end{tabular}
  \caption{Overall and type-wise action prediction performance on AITZ when training Qwen2-VL-7B with grounding, query inference as the alternative task, and query inference as a pivot task, combined with different CoAT components. The optimal and the suboptimal results are \textbf{bolded} and \underline{underlined}, respectively. }
  \label{tab:aitzAblation}
\end{table*}

Recently, the success of CoT in large scale of MLLMs~\citep{chu2024navigate} has flourished its widely deployment. To thoroughly investigate the influence of CoT for 7B-level perception-enhanced MLLMs in resource-constrained scenarios, we adopt the chain-of-action-thought (CoAT) dataset AITZ~\citep{zhang2024android} to subsequently fine-tune perception-enhanced  MLLMs and access their respective action prediction performance with different CoAT components. The overall and type-wise results are presented in Table~\ref{tab:aitzAblation}.

Specifically, we utilize four components of CoAT: screen descriptions (SD) and previous action results (PAR) as additional semantic information in inputs, along with action thoughts (AT) and next action descriptions (AD) as intermediate reasoning results in outputs. To examine the influence of both input and output components, we categorize the experiments into four groups: (i) without any CoAT components (ID 1 in Table~\ref{tab:aitzAblation}); (ii) only with input components (ID 2–4 in Table~\ref{tab:aitzAblation}); (iii) only with output components (ID 5–7 in Table~\ref{tab:aitzAblation}); and (iv) combining both input and output components (ID 8–10 in Table~\ref{tab:aitzAblation}). Based on the results, we have the following findings:

(i) Generally, incorporating additional semantic information into inputs further improves action prediction performance. For example, when combining PAR with query inference as the alternative task, the AMR reaches 67.06, while combining SD with query inference as the pivot task results in an AMR of 67.27, both surpassing the 67.04 achieved by Atlas, as presented in Table~\ref{tab:reasoningImprovements}. Additionally, grounding-enhanced models also benefit from the additional semantic information in inputs, leading to further improvements in action prediction.  These observations indicate that providing additional semantic information to inputs enhances the perception of GUI environments, ultimately leading to more accurate action decisions.

(ii) Incorporating intermediate reasoning results to outputs yields significant degradation in action prediction performance. For instance, when combining AT with query inference as the pivot task, AMR drops to 61.39, which is substantially lower than the performance without CoAT components.  The degradation becomes even more pronounced when both input and output components are included, with the AMR falling below 60\%. This decline may be attributed to the relatively small scale of the 7B model, which struggles to process complex reasoning effectively. When lengthy intermediate reasoning results are introduced, the model may become overly focused on the reasoning chain itself rather than the final action decision.

(iii) Adopting query inference generally outperforms grounding when combined with different CoAT components. Within each group of the same ID, adopting query inference either as alternative task or pivot task generally outperforms grounding, highlighting the effectiveness of query inference when combined with CoT-enhanced reasoning.

In summary, incorporating additional semantic information into inputs for query inference further enhances reasoning performance, offering an alternative path for improving action prediction in resource-constrained scenarios.

\section{Conclusions}\label{sec:conclusions}
In this paper, we identify the performance gap between coordinate-oriented grounding and action-oriented reasoning in resource-constrained scenarios.  To smooth grounding and reasoning, we propose query inference, a query-oriented approach designed to enhance the comprehension of user intent while maintaining sensitivity to grounding coordinates. Experimental results demonstrate that query inference outperforms grounding under same data scale. Notably, query inference achieves performance comparable to large-scale grounding-enhanced OS-Atlas with significantly less training data. Additionally, incorporating additional semantic information into inputs for query inference provides an alternative approach to further improving action prediction in resource-constrained scenarios.

\section*{Limitations}
Our approach has limitations in two main aspects. First, our method focus on enhancing perception for reasoning with a small-scale dataset, which may weaken the zero-shot capability of the MLLM, thereby requiring SFT on specific agent benchmarks. Second, as we only focus on resource-constrained scenarios, the results may differ with large-scale training data, as grounding has been shown to be effective in such settings.

\section*{Ethics Statement}
This section outlines the ethics considerations in the following aspects:  (i) Privacy. The research dataset UIBERT~\citep{bai2021uibert} is a publicly available dataset that extended from the public Rico dataset~\citep{deka2017rico}, containing no toxic, biased, misleading content, or personal privacy. The two mobile agent benchmarks, AndroidControl~\citep{li2024effects} and AITZ~\citep{zhang2024android} are all publicly available datasets which also implemented safeguards protect privacy. Moreover, we provide an approach to bridge grounding and reasoning in resource-constrained scenarios and support local deployment. (ii) System security. As we train MLLMs to act as the brain of GUI agents, emulating human-like behaviors, security measures are better aligned with human-oriented mechanisms, which are already integrated into existing GUI systems for operating systems. (iii) Potential social impacts. Our proposed query inference can further improve reasoning performance of GUI agents in resource-constrained scenarios. However, malicious actors may exploit GUI agents for harmful purposes. To mitigate the risks, platforms may need to update detection, authorization, and governance protocols to address potential social implications.

\bibliography{arxiv.bib}

\appendix

\begin{figure*}[!t]
  \centering
  \includegraphics[width=\linewidth]{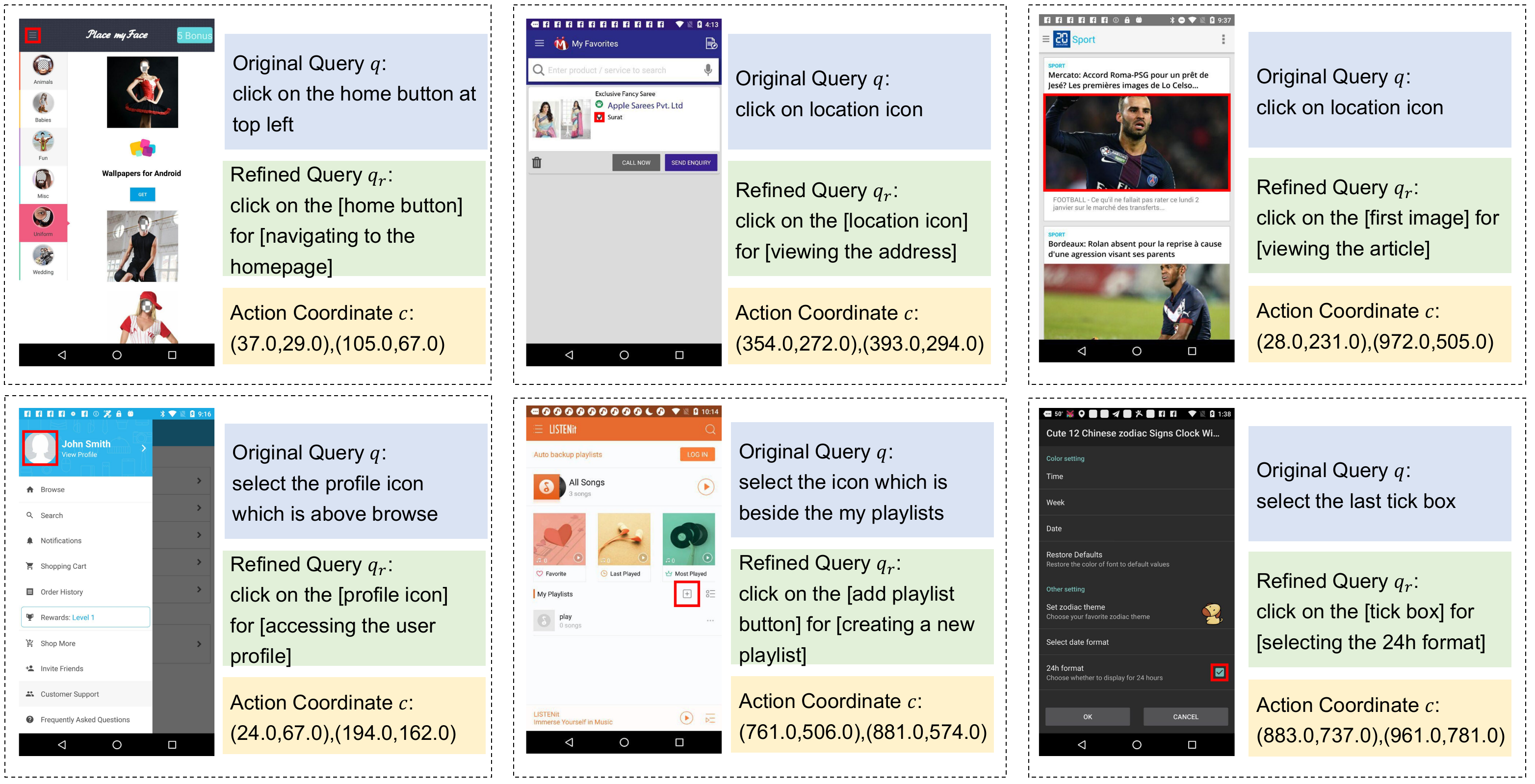}	
  \caption{Example triplets $\langle s, q_r, c \rangle$ from the refined UIBERT dataset, along with the original query $q$.}
  \label{fig:refinedUIBERT}
\end{figure*}

\begin{table*}[!t]
  \centering
  \small
  \setlength{\tabcolsep}{8pt}
    \begin{tabular}{@{}cccccccccc@{}}
      \toprule
      Dataset        & \texttt{SCROLL} & \texttt{CLICK} & \texttt{TYPE} & \texttt{PRESS} & \texttt{WAIT} & \texttt{OPENAPP} & \texttt{COMPLETE} & Others & Total \\ \midrule
      AndroidControl & 1,211  & 5,074 & 632  & 343   & 567  & 608     & 1543     & 9      & 9,987 \\ 
      AITZ           & 601    & 2,736 & 500  & 265   & /    & /       & 504      & 118    & 4724  \\ \bottomrule
      \end{tabular}
  \caption{Action type distributions of AndroidControl and AITZ test subset.}
  \label{tab:datasetDetail}
\end{table*}
\section{Detailed Experimental Setup} \label{appendix:detailedSetup}
This section presents additional setup information for the experiments. Section~\ref{subappendix:refinedUIBERT} presents examples from the refined UIBERT dataset for query inference. Section~\ref{subappendix:datasets} details the AndroidControl and AITZ benchmarks. Section~\ref{subappendix:amr} outlines the evaluation process for AMR. Section~\ref{subappendix:usageArtifacts} discusses the usage of existing artifacts.

\subsection{Examples of Refined UIBERT} \label{subappendix:refinedUIBERT}

Example triplets $\langle s, q_r, c \rangle$ from the refined UIBERT dataset, along with the original query $q$ are provided in Figure~\ref{fig:refinedUIBERT}. After refinement, the action intent has been inferred, such as ``\textit{selecting the 24h format}''. By training on the triplets $\langle s, q_r, c \rangle$ with intended queries, the comprehension of user intention would be enhanced to align with reasoning while maintaining sensitivity to the coordinates.

\subsection{Details of AndroidControl and ATIZ} \label{subappendix:datasets}
The details of the AndroidControl and AITZ datasets are as follows:

\noindent $\bullet$ AndroidControl~\citep{li2024effects} is a mobile agent dataset comprising 15,283 demonstrations with step-wise instructions. This dataset is collected from human raters performing various tasks on 833 different apps spanning 40 app categories on Android devices. The training subset of AndroidControl includes 89,144 step-wise samples.

\noindent $\bullet$ AITZ~\citep{zhang2024android} is a mobile agent dataset derived from a subset of AITW~\citep{rawles2023android} and annotated by proprietary MLLMs for CoAT components. AITZ consists of 2,504 operation trajectories across 18,643 steps. AITZ is categorized into five subsets based on application domain: General, Install, GoogleApps, Single, and Web Shopping. The training subset of AndroidControl contains 13,919 step-wise samples.

The action type distributions of the AndroidControl and AITZ test subsets are presented in Table~\ref{tab:datasetDetail}.

\subsection{Evaluation of AMR} \label{subappendix:amr}
The exact action match rate (AMR) is a more accurate metric for evaluating step-wise action prediction. AMR considers both the action type $t$ and optional parameters $p$ (e.g., coordinates, app name, text input). An action is considered an exact match only when both $t$ and $p$ align perfectly with the ground truth.  The calculation of AMR varies depending on the action type, as outlined below:

For action without additional parameters, including \texttt{WAIT}, \texttt{COMPLETE}, and \texttt{PRESS}, we focus solely on matching the action type between predicted actions and the ground truth. AMR is equivalent to TMR for these actions.

For \texttt{SCROLL} actions, where the direction can only be up, down, left, or right, we evaluate both the action type and the scroll direction to ensure they perfectly align with the ground truth.

For text-based actions, including \texttt{TYPE} and \texttt{OPENAPP}, we adopt a \textbf{rigorous examination}, where the predicted action is considered an exact match only when both the action type and corresponding text (e.g., typed content and app names) perfectly align with the ground truth.

For \texttt{CLICK} actions, as both AndroidControl and AITZ datasets provide the layout information of the screenshots, we adapt the evaluation method from~\citet{wu2025osatlas}. Specifically, when both the predicted and ground truth actions are \texttt{CLICK}, we first examine the corresponding screenshot layout to locate the element bounding box that contains the ground truth coordinates. If a bounding box is found, we check whether the predicted coordinates fall within it. If so, the \texttt{CLICK} action is deemed correctly predicted; otherwise, it is not. If no bounding box is found, we compute the relative distance between the predicted and ground truth coordinates, considering the \texttt{CLICK} action correct if the relative distance is below 14\% of the screen. 

\subsection{Usage of Existing Artifacts} \label{subappendix:usageArtifacts}
We adopt LLaMa-Factory~\citep{zheng2024llamafactory} for grounding and query inference and SFT on mobile agent benchmarks. Besides, we adopt Huggingface transformers\footnote{\url{https://github.com/huggingface/transformers}} to load MLLMs for testing. For acceleration during testing, we employ flash-attn~\citep{dao2023flashattention2}. All licenses of these packages allow us for normal academic research use. All experiments are conducted on $4 \times $ NVIDIA A100, each with 80GB GPU memory. Training for query inference and grounding takes approximately 2 hours. Fine-tuning on AITZ also requires about 2 hours, whereas fine-tuning on AndroidControl takes approximately 14 hours.

\section{Prompts}\label{appendix:prompts}
\begin{figure}
  \centering
  \includegraphics[width=\linewidth]{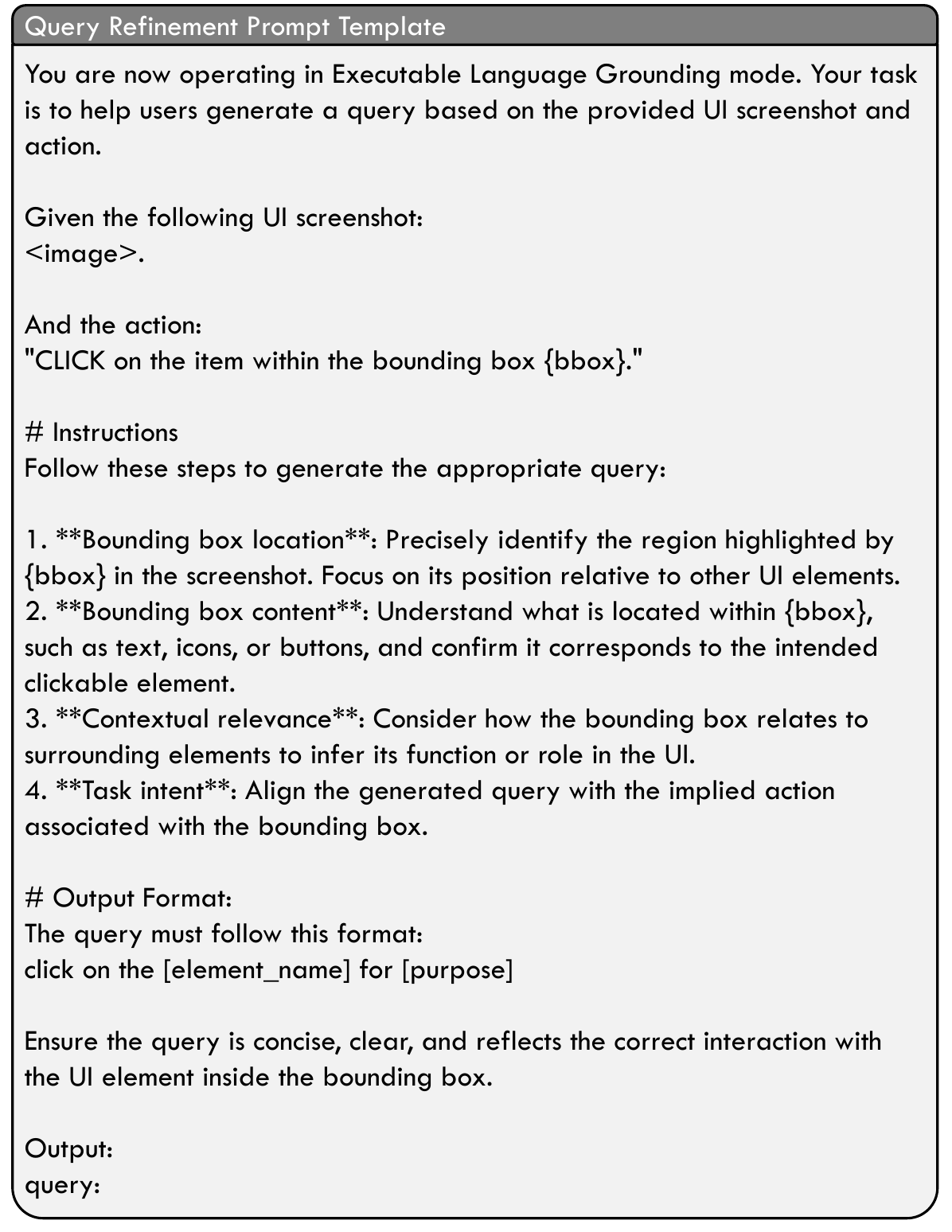}
  \caption{The prompt template for query refinement. }
  \label{prompt:queryRefinement}
\end{figure}

\begin{figure}
  \centering
  \includegraphics[width=\linewidth]{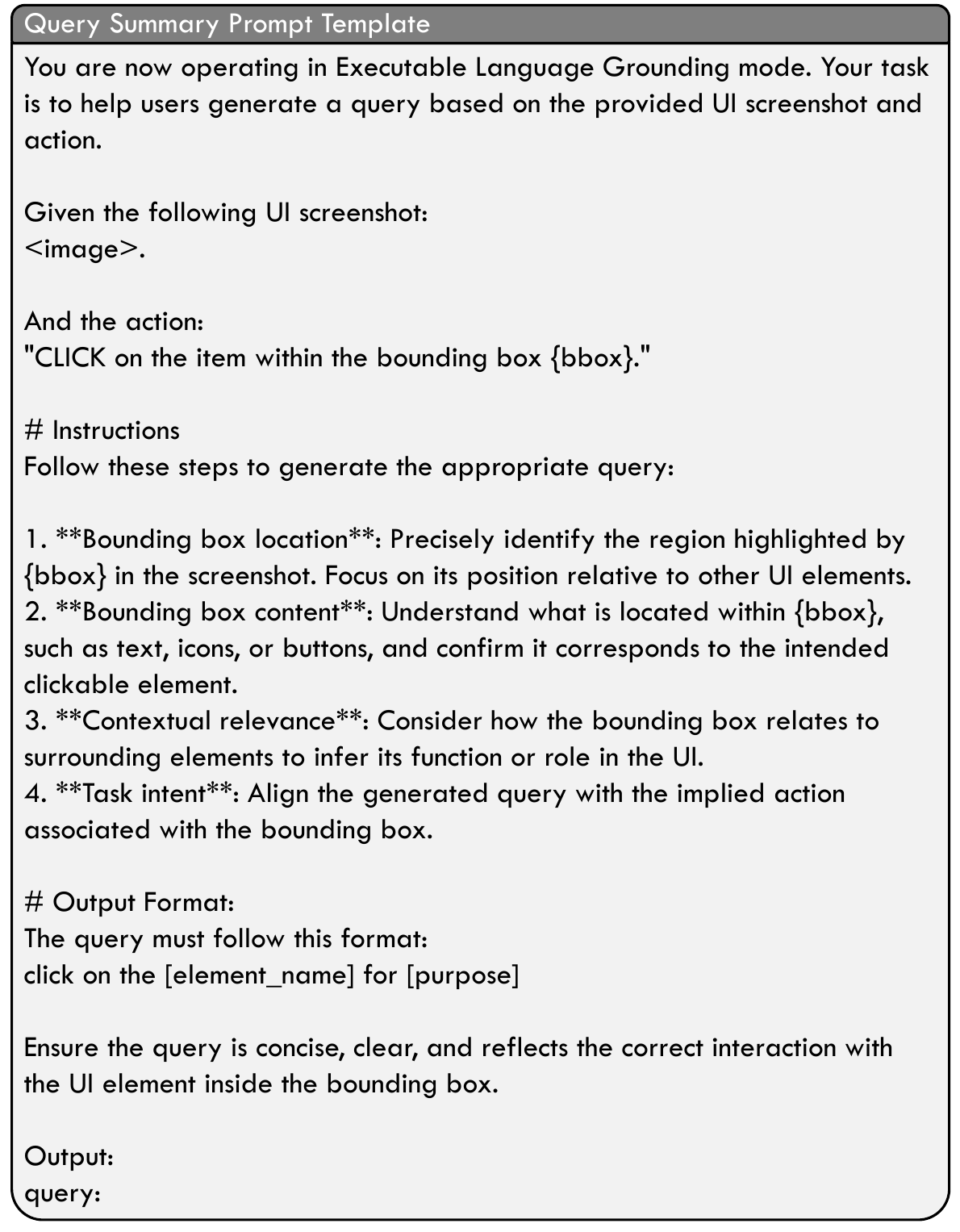}
  \caption{The prompt template for query inference. }
  \label{prompt:querysummary}
\end{figure}

\begin{figure}
  \centering
  \includegraphics[width=\linewidth]{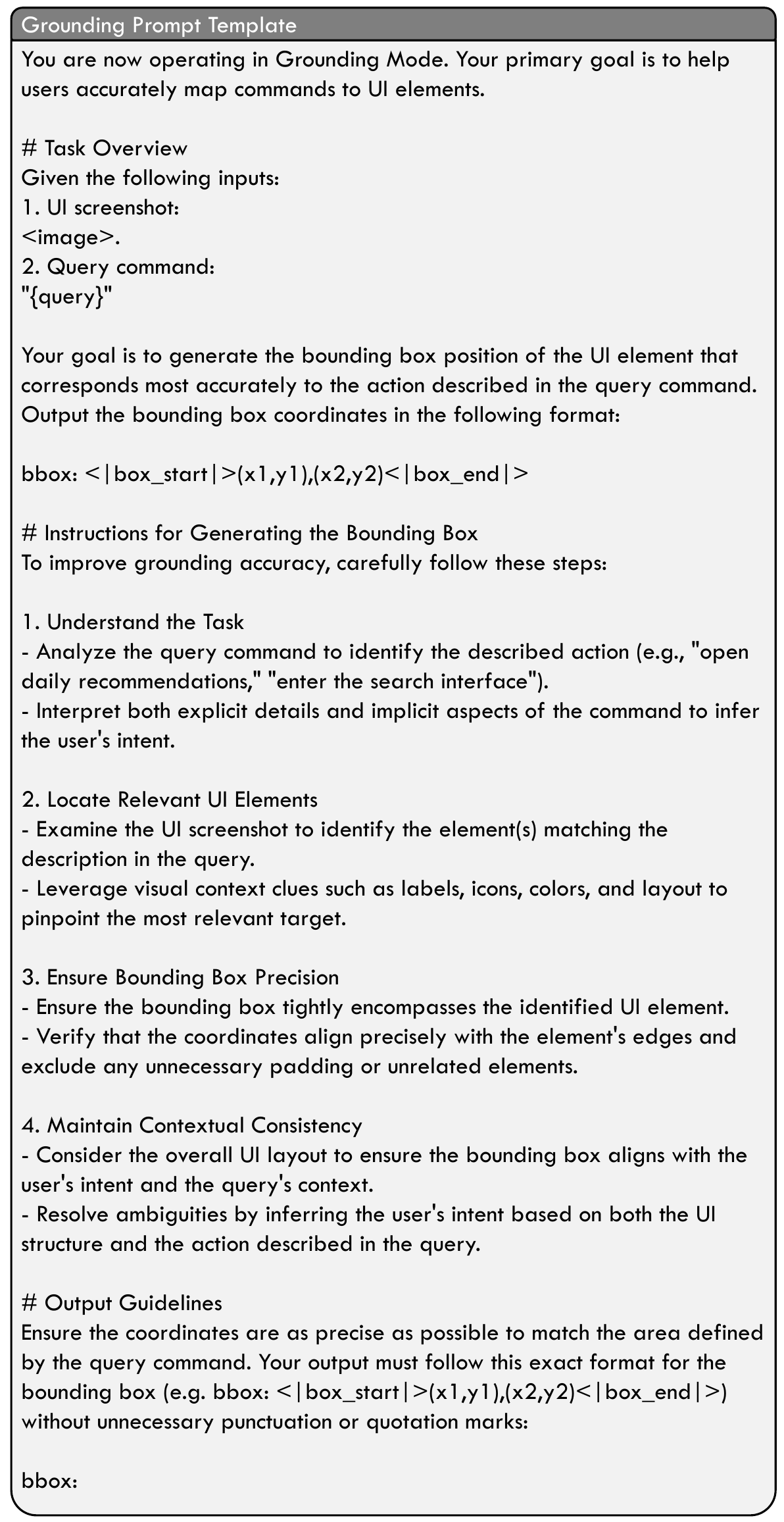}
  \caption{The prompt template for grounding. }
  \label{prompt:grounding}
\end{figure}

\begin{figure}
  \centering
  \includegraphics[width=\linewidth]{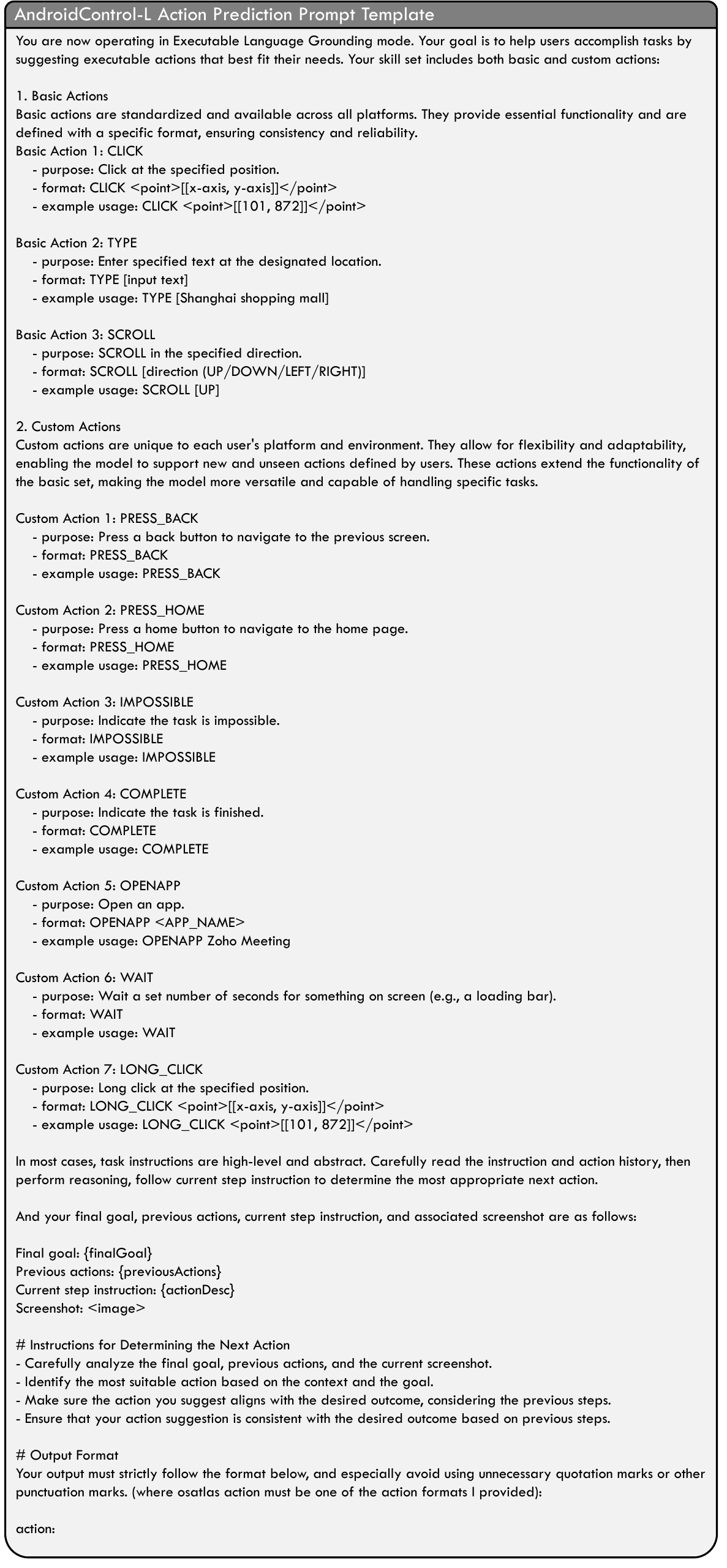}
  \caption{The prompt template for action prediction on AndroidControl-L. }
  \label{prompt:AndroidControl-L}
\end{figure}

\begin{figure}
  \centering
  \includegraphics[width=\linewidth]{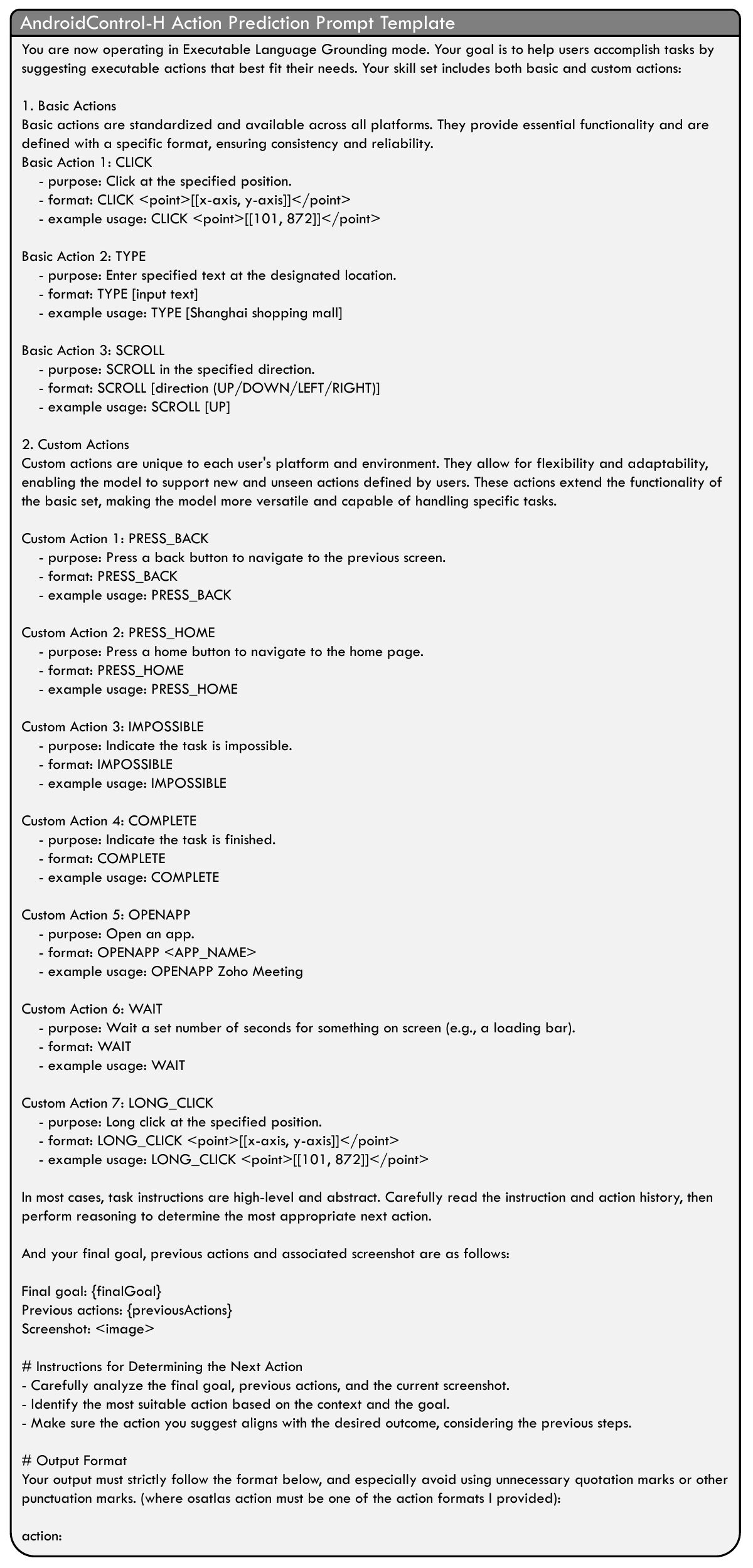}
  \caption{The prompt template for action prediction on AndroidControl-H. }
  \label{prompt:AndroidControl-H}
\end{figure}

\begin{figure}
  \centering
  \includegraphics[width=\linewidth]{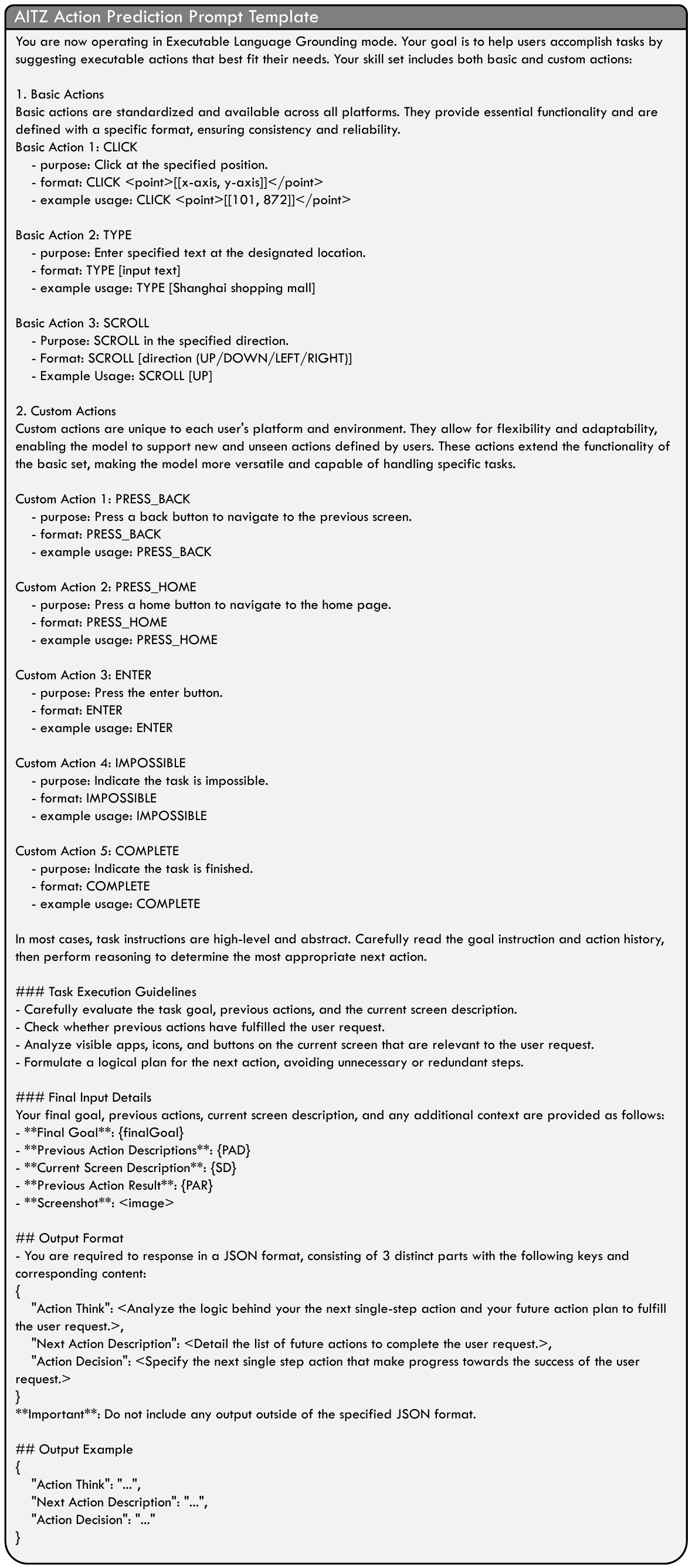}
  \caption{The prompt template for action prediction on AITZ with SD, PAR, AT, and AD. }
  \label{prompt:AITZ}
\end{figure}

This section presents our meticulously designed prompts. Specifically, for constructing the query inference dataset, the prompt template for query refinement is shown in Figure~\ref{prompt:queryRefinement}. For grounding and query inference, the prompt templates are presented in Figure~\ref{prompt:grounding} and Figure~\ref{prompt:querysummary}, respectively. For reasoning, prompt templates for AndroidControl-L and AndroidControl-H are provided in Figures~\ref{prompt:AndroidControl-L} and Figure~\ref{prompt:AndroidControl-H}, respectively. Additionally, the prompt template for AITZ, when combined with SD, PAR, AT, and AD, is provided in Figure~\ref{prompt:AITZ}.

\end{document}